\documentclass{sig-alternate}

\usepackage{multirow}
\usepackage{epstopdf}
\usepackage{pgfplots}
\usepackage{algorithmic}
\usepackage{algorithm}
\usepackage{subfigure}
\usepackage{amsmath}
\usepackage{amssymb}  

\begin{document}
%
\conferenceinfo{WOODSTOCK}{'97 El Paso, Texas USA}

\title{Learning Better Encoding for Approximate Nearest Neighbor Search with Dictionary Annealing}

\numberofauthors{2}
\author{
	Liu Shicong, Lu Hongtao
}

\maketitle
\begin{abstract}
We introduce a novel dictionary optimization method for high-dimensional vector quantization employed in approximate nearest neighbor (ANN) search. Vector quantization methods first seek a series of dictionaries, then approximate each vector by a sum of elements selected from these dictionaries. An optimal series of dictionaries should be mutually independent, and each dictionary should generate a balanced encoding for the target dataset. Existing methods did not explicitly consider this. To achieve these goals along with minimizing the quantization error (residue), we propose a novel dictionary optimization method called \emph{Dictionary Annealing} that alternatively "heats up" a single dictionary by generating an intermediate dataset with residual vectors, "cools down" the dictionary by fitting the intermediate dataset, then extracts the new residual vectors for the next iteration. Better codes can be learned by DA for the ANN search tasks. DA is easily implemented on GPU to utilize the latest computing technology, and can easily extended to an online dictionary learning scheme. We show by experiments that our optimized dictionaries substantially reduce the overall quantization error. Jointly used with residual vector quantization, our optimized dictionaries lead to a better approximate nearest neighbor search performance compared to the state-of-the-art methods.
\end{abstract}


\section{Introduction}
\begin{figure}
\begin{center}
   \includegraphics[width=0.8\linewidth]{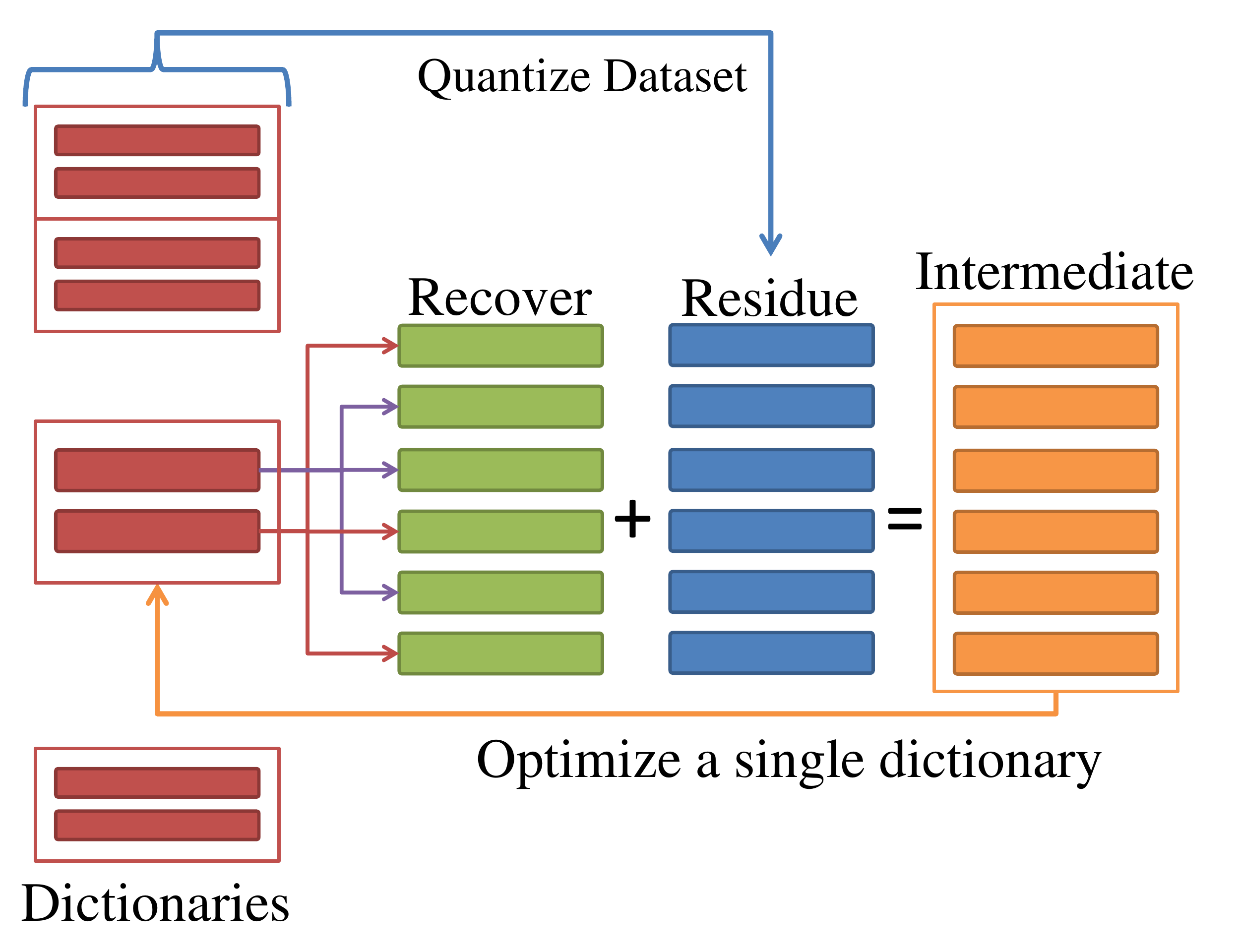}
\end{center}
   \caption{On each iteration, Dictionary Annealing first picks a dictionary, and generates an intermediate dataset with the residue and the dictionary, then optimizes the picked dictionary to better fit the intermediate dataset, finally quantizes the dataset to obtain the residue for next iteration. The figure is best viewed in color.}
\label{figIllustration}
\end{figure}

Since the seminal work of Product Quantization(PQ)\cite{pq}, there has been a growing interest in the computer vision community to apply vector quantization to high-dimensional large scale dataset before any applications, to fight the curse of dimensionality\cite{curse}. A typical scenario is approximate nearest neighbor(ANN) search task, which has been a fundamental problem in many computer vision applications such as image retrieval \cite{IR1} and image recognition \cite{IR2}. Traditional ANN search methods include hashing based methods Locality Sensitive Hashing\cite{lsh}, Iterative Quantization\cite{itq}, Spectral Hashing\cite{SH}, Kernelized Locality Sensitive Hashing \cite{ker}, LDAHash\cite{lhs}, etc, they transform an original database vector into a sequence of bits, and then use hamming distances to approximate the distances between vectors in the embedded hashing codes. Data structures such as Hierarchical K-means\cite{hkm}, KD-Tree\cite{kd}, R-Tree\cite{RTree}, X-Tree\cite{XTree} are also proposed to perform ANN tasks.

Product Quantization \cite{pq} is a novel vector quantization method for nearest neighbor search. PQ divides the feature space into $M$ disjoint subspaces of same dimensions, and performs k-means to learn $M$ dictionaries with $K$ elements per dictionary on these lower-dimensional subspaces. Then the original database vectors are approximated with the concatenation codings of $M$ elements chosen one per dictionary. PQ and its variations allow fast distance computation to perform efficient ANN search. Given a query vector $q$, the distances between $q$ and each element from the dictionaries are precomputed. Then the distances to other database vectors can be efficiently approximated by $M$ lookup tables. Thus a linear scan procedure could be accelerated hundred-fold by PQ. Compared to hashing methods, the search accuracy of PQ is much higher within the same search time\cite{arandjelovic2012multiple}.

To further improve the performance of PQ, optimized product quantization(OPQ) \cite{opq} and Cartesian k-means (ck-means)\cite{ck} find an optimized rotation for better subspace partition and further lower the quantization error of PQ. Composite Quantization\cite{cq} and Additive Quantization\cite{babenko2014additive} generalize PQ by relaxing the constraint of PQ that decomposed data space into orthogonal subspaces. Distance-encoded product quantization \cite{heo2014distance} extends PQ by encoding both cluster index and the distance to the cluster center. However, these methods mainly focus on relaxing constraints or introducing new parameters to improve PQ. How to incrementally improve the dictionaries learned initially so as to further improve vector quantization performance remains largely un-addressed.

For hashing-based binary embedding methods, for example. Spectral Hashing\cite{SH}, Semi-supervised Hashing\cite{wang2010semi}, they aim to find an efficient code that each bit has a 50 \% chance of being 1 or 0, and that different bits are independent of each other. Similarly, for quantization-based embeddings, which encode an original vector into several chunks, we aim to find an encoding that each chunk has a $1/K$ chance of being ${1\cdots K}$, and that different chunks are independent of each other. That means for each dictionary, elements should be evenly chosen by database vectors, also dictionaries should be mutually independent. Among all dictionaries meeting these requirements, we seek the one makes the quantization error minimal.

In this paper we propose a new dictionary optimization method called \emph{Dictionary Annealing} (DA) which alternatively optimizes a single dictionary with residue and re-encode the dataset to obtain the latest residue. See Figure \ref{figIllustration} for an intuitive depiction of DA algorithm. 
Inspired by simulated annealing, the main idea of DA is to "heat up" a dictionary to a better initial position so we can "cool down" the dictionary with smaller residue left.
Given a series of learned dictionaries by a quantization method, say, Residue Vector Quantization, on each iteration, DA
\begin{enumerate}
\item Sorts the dictionaries by their elements' norm, then uses a beam search method to fit the dataset with the dictionaries and obtain the residue;
\item Picks a single dictionary to optimize: first generates an intermediate dataset by the sum of the residue and the components of the quantized dataset on this dictionary, then optimizes this dictionary to better fit the intermediate dataset. 
\end{enumerate}
Similar to subspace clustering presented in \cite{cheng1999entropy}, DA incrementally optimizes a single dictionary via subspaces. DA first performs k-means on a $d'$-dimensional subspace (where $d'$ depends on the information entropy of this dictionary) initialized by the dictionary elements on this subspace, then iteratively adds more dimensions and performs k-means on this higher-dimensional subspace, initialized by the optimized dictionary on the previous iteration (elements padded with zeros chunks). This process is repeated until we have fitted the whole feature space. 

Our proposed \emph{Dictionary Annealing} is closely related to the Residual Vector Quantization(RVQ) \cite{rvq}, which generates mutually independent dictionaries by directly quantizing on the residual vectors. However, the performance of RVQ is limited by unbalanced partition for the later stages of quantization. Nevertheless, the residual vectors can be used to increase the independence of dictionaries and the unbalanced partition problem can be solved via initialization on subspace. The empirical results show that our \emph{Dictionary Annealing} indeed finds a better encoding. We have validated our methods on two commonly used datasets for evaluating ANN search performance: SIFT-1M and GIST-1M \cite{pq}. The dictionaries optimized by our method gained a significant performance boost compared to other un-optimized state-of-the-art methods.

In addition, our algorithm could be easily applied to online dictionary learning. For ANN tasks, the major concern is to speed up the query process while maintaining a high precision and recall, while it's acceptable to spend more time on dictionary learning and encoding. In our algorithm, the dictionaries learned previously are not discarded but improved, so our online dictionary learning is done simply by feeding new-coming data in big batches. Online dictionary learning for matrix factorization and sparse coding has been proposed in \cite{mairal2009online}, while our algorithm aims to boost performance of ANN tasks. Experiments show that our online dictionary learning substantially further improves the ANN search quality, which makes vector quantization methods more effective to the ever-growing dataset in the real world applications.

The remainder of this paper is organized as follows: We first briefly introduce quantization methods for ANN tasks in Section \ref{secRW}. In Section \ref{secEncoding}, we briefly discuss what makes good encoding for quantization methods, and present the observation on popular quantization methods. In Section \ref{secDA}, we propose our \emph{Dictionary Annealing} algorithm. In Section \ref{secDis}, we discussed the initialization, scalability and implementation of \emph{Dictionary Annealing}. Finally we evaluate our method for ANN tasks, and compared to other state-of-the-art quantization methods to demonstrate the superiority of the optimized dictionaries learned by \emph{Dictionary Annealing}.

\section{Quantization for ANN Search}
\label{secRW}
The main advantage of quantization method for approximate nearest neighbor search is that Asymmetric Distance Computation (ADC) introduced in \cite{pq} allows fast and accurate distance approximation. Denote any $\mathbf{x}$ in dataset $\mathbf{X}$, with ADC, we can exhaustively compute the distance between a query vector $\mathbf{q}$ and all the vectors $\mathbf{x}\in\mathbf{X}$. Quantization methods for ANN search use a series of, say $M$ dictionaries $\mathbf{C}_m=\{\mathbf{c}_m(1),\cdots, \mathbf{c}_m(K)\}, m=1,\cdots,M$, each containing $K$ elements, to approximate a database vector as the sum of $M$ vectors sequentially chosen from these dictionaries: 

$$ \mathbf{x} \approx \sum_{m=1}^M \mathbf{c}_m(i_m(\mathbf{x})),$$
where $i_m(\mathbf{x})$ is the index function of $\mathbf{x}$. Then the Euclidean distance between an input query $\mathbf{q}$ and a database vector $\mathbf{x}$ is approximated by:

\begin{equation}
\label{appr}
\begin{split}
\lVert\mathbf{q}-\mathbf{x}\rVert^2&\approx\lVert \mathbf{q}-\sum_{m=1}^M \mathbf{c}_m(i_m(\mathbf{x}))\rVert^2\\ 
&=\sum_{m=1}^M\lVert \mathbf{q}-\mathbf{c}_m(i_m(\mathbf{x}))\rVert^2-(m-1)\lVert \mathbf{q}\rVert^2 \\
&\quad+\sum_{i=1}^M\sum_{j=1,j\neq i}^M \mathbf{c}_i(i_i(\mathbf{x}))^\mathrm{T}\mathbf{c}_j(i_j(\mathbf{x}))
\end{split}
\end{equation}

For every query $\mathbf{q}$, the first term is precomputed before the exhaustive distance computation, the second term is a constant for all database vectors which can be omitted, and the third term is precomputed on database encoding stage. Thus, the approximate distance between $\mathbf{q}$ and a database vector $\mathbf{x}$ can be efficiently computed in $M$ table lookups and $M$ addition.

Product Quantization generates dictionaries on the disjoint subspaces, so the requirement of computing the third term is eliminated. Composite Quantization \cite{cq} introduced an  inner-dictionary-element-product to put constraint on the third term above, and the need for computing this term is also eliminated. Additive Quantization \cite{babenko2014additive} and Residual Vector Quantization \cite{rvq} require the third term to be encoded together with the dataset to perform the ADC. 

\section{Good encoding for quantization methods}
\label{secEncoding}

\begin{figure*}[t]
\begin{center}
	\subfigure[Additive Quantization]{
    	\includegraphics[width=0.3\linewidth]{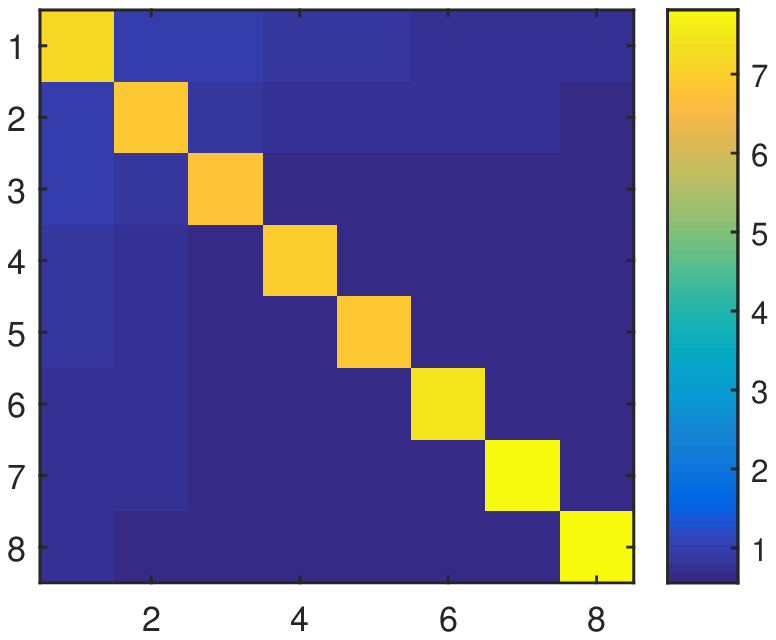}
    }
	\subfigure[Product Quantization]{
    	\includegraphics[width=0.3\linewidth]{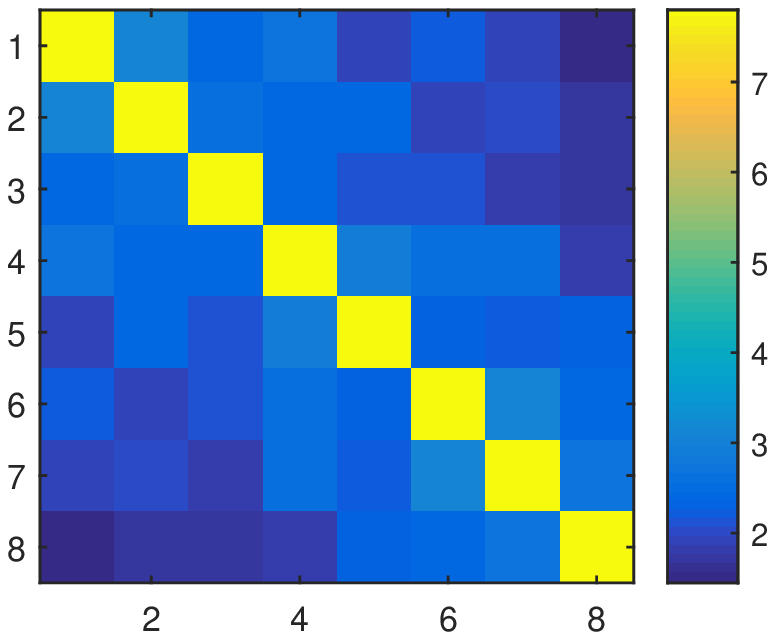}
    }
	\subfigure[Optimized Product Quantization]{
    	\includegraphics[width=0.3\linewidth]{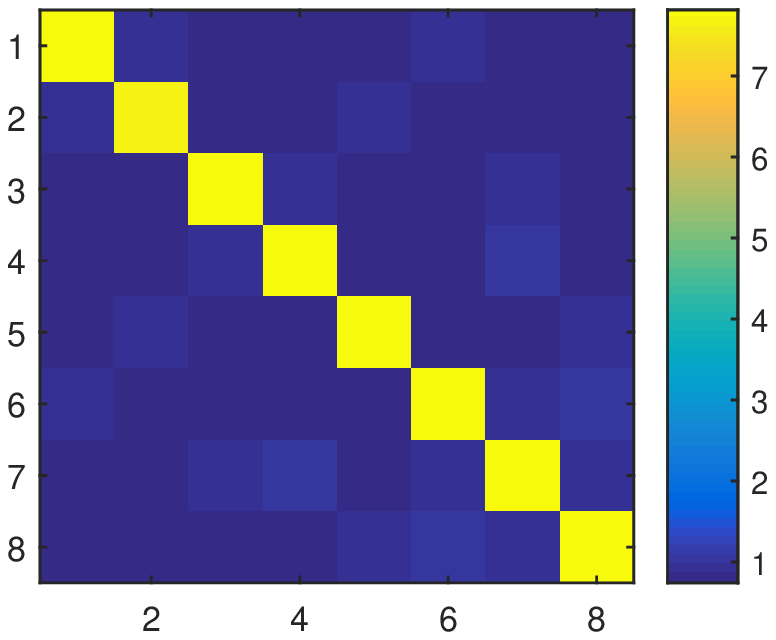}
    }
	\subfigure[Residual Vector Quantization]{
    	\includegraphics[width=0.3\linewidth]{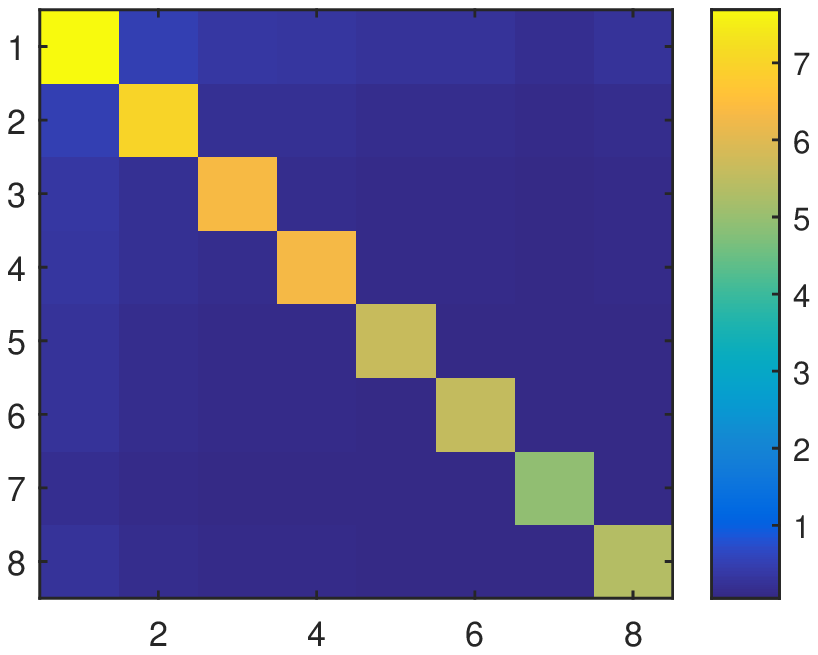}
    }
	\subfigure[Dictionary Annealing, warm started by dictionaries learned with RVQ]{
    	\includegraphics[width=0.3\linewidth]{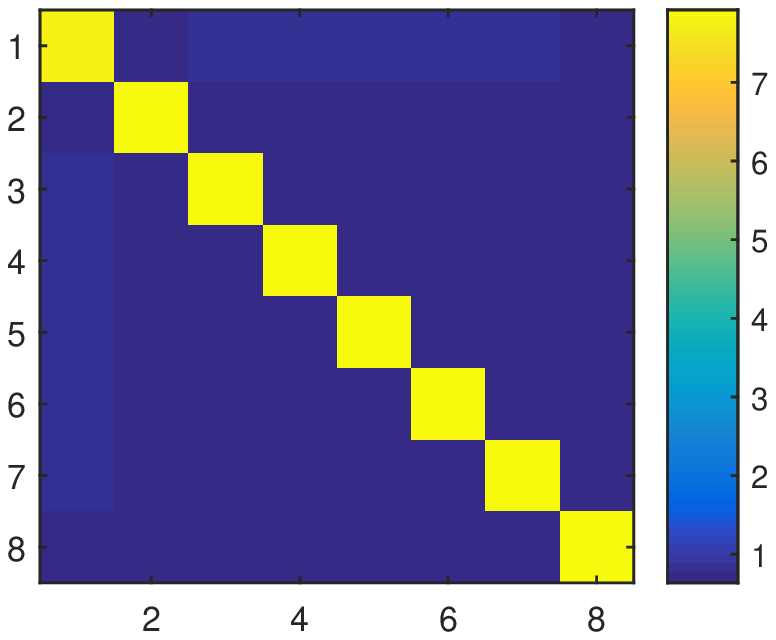}
    }           
    \subfigure[Information Entropy of dictionaries learned by different methods]{
	
	\pgfplotsset{every axis/.append style={
                    legend style={font=\tiny},
                    }}
   \pgfplotscreateplotcyclelist{my}{
   	{blue, mark=*},
   	{dashed, red,mark=square,mark options={solid}},
   	{dashed, violet,mark=o,mark options={solid}},
   	{dashed, cyan,mark=+,mark options={solid}},
   	{teal, mark=o},
   	{yellow!40!red, mark=triangle},
   	{magenta, mark=diamond},
   	{black!20!green, mark=otimes},
   }
   	\begin{tikzpicture}
   		\begin{axis}[
   			width=0.3\textwidth,
   			height=0.22\textwidth,
   			axis y line=left,
   			axis x line=bottom,
   			ylabel=Information Entropy,
   			ylabel near ticks,
   			xlabel=Dictionary,
   			xtick={1,2,3,4,5,6,7,8},
   			xticklabels={1,2,3,4,5,6,7,8},
   			ytick={4,5,6,7,8},   			
   			ymin=4, ymax=8,
   			legend pos=south west,
   			cycle list name=my,
   			grid,
   			]
   			
   			\addplot table [x=D, y=PQ, col sep=comma] {figures/mutualInfo/entropy.txt};

   			\addplot table [x=D, y=OPQ, col sep=comma] {figures/mutualInfo/entropy.txt};

   			\addplot table [x=D, y=AQ, col sep=comma] {figures/mutualInfo/entropy.txt};

   			\addplot table [x=D, y=RVQ, col sep=comma] {figures/mutualInfo/entropy.txt};

   			\addplot table [x=D, y=DA, col sep=comma] {figures/mutualInfo/entropy.txt};
   			
   			\legend{PQ,OPQ,AQ,RVQ,DA};
   		\end{axis}
   		
   		\end{tikzpicture}
    }
\caption{Mutual Information Matrix between dictionaries for different quantization methods. Experiment conducted on a subset containing 100K 960-d vectors from GIST-1M dataset. We used different methods to learn $M=8$ dictionaries, $K=256$ elements per dictionary. The perfect encoding should have no mutual information between different dictionary and has an information entropy of $\log K=8$bits for each dictionary. Our proposed method achieves near optimal encoding.}
  \end{center}
\label{figInfo}

\end{figure*}

For hashing based approximate nearest neighbor search methods, we seek a code that only requires a small number of bits to represent the full dataset while maps similar items to similar binary codewords. An efficient code requires that each bit has a 50\% chance of being one or zero, and different bits are mutually independent. This is usually done by thresholding and find optimal orthogonal projections like in Spectral Hashing\cite{SH}, Iterative Quantization\cite{itq}, Semi-supervised Hashing\cite{wang2010semi}, etc. 

For quantization based approximate nearest neighbor search methods, the criterion for efficient code is essentially the same as the hashing based methods. We would like to obtain maximum information entropy($S(\mathbf{C}_m)$) for every dictionary $\mathbf{C}_m$ and no mutual information between different dictionaries: 

\begin{equation}
\begin{split}
S(\mathbf{C}_m)=\sum_{k=1}^K p_k^m(\log_2 p_k^m)&=\log_2K \\
\sum_{k_i,k_j\in 1\cdots K} p_{ij}(k_i,k_j)&\log_2\frac{p_{ij}(k_i,k_j)}{p_{k_i}^i p_{k_j}^j}=0\\
& for\quad i,j\in 1\cdots M
\end{split}
\end{equation}

where $p_k^m$ denotes the probability of dictionary that in $\mathbf{C}_m$, $k$-th element is chosen; and $p_{ij}(k_i,k_j)$ denotes the probability that $k_i$-th element from $\mathbf{C}_i$ and $k_j$-th element from $\mathbf{C}_j$ is chosen by a vector $\mathbf{x}$ simultaneously. We present an illustrative comparison of encoding quality with the criterion above between different quantization methods in Figure \ref{figInfo}.

To obtain balanced partitions, PQ clusters on disjoint subspaces, however these subspaces could be correlated. To obtain independent dictionaries, previous works pre-process the data using simple heuristics like randomly ordering the dimensions \cite{pq} or randomly rotating the space \cite{jegou2010aggregating}. Optimized product quantization and Cartesian k-means further find an optimal rotation of original feature space so that dimensions are de-correlated.

Residual vector quantization(RVQ)\cite{rvq} uses a different approach to obtain mutually independent dictionary simply by learning dictionaries on the residual brought by the previously learned dictionaries. However RVQ suffers from less efficient single dictionary, because k-means is not really meant for clustering on high-dimensional data as depicted in \cite{steinbach2004challenges}. K-means algorithm fails to generate good quality dictionary on the residual spaces, a direct observation is the low information entropy on the latter dictionaries.

The final goal of a good encoding is to lower the quantization error(the residue):

$$E(\mathbf{C}_1,\mathbf{C}_2,\cdots,\mathbf{C}_M)=\sum _{\mathbf{x}\in\mathbf{X}}\lVert \mathbf{x}-\sum_{m=1}^M \mathbf{c}_m(i_m(\mathbf{x}))\rVert^2$$

Given a series of learned dictionaries, though they may encode the dataset not so well, they still contain much information on the structure of the dataset. \emph{Dictionary Annealing} seeks an incremental refinement to such dictionaries.

\section{Dictionary Annealing}
\label{secDA}
The main idea of our proposed Dictionary Annealing  is to use residual vectors to generate an intermediate dataset, i.e, "heating up" dictionary. Then "cools down" the dictionary by fitting the intermediate dataset. We have two reasons for doing so:
\begin{itemize}
\item The residual space are largely independent to other dictionary spaces, as observed in Figure \ref{figInfo}. If a dictionary fits the residual space well, then it gains much independence.

\item The intermediate dataset is actually part of the original dataset. So if a dictionary fits the intermediate dataset better, then the quantization error is also reduced.
\end{itemize}
DA also manages to find a balanced partition, we'll explain it in the following texts. See Algorithm \ref{algIOlearn} for a brief pseudo code for Dictionary Annealing. 

\begin{algorithm}

\caption{Dictionary Annealing}
\label{algIOlearn}
\textbf{Input}: Dataset $\mathbf{X}$, dimensions $d$, number of dictionaries $M$, initial dictionaries $\{\mathbf{C}_m, m\in1\cdots M\}$, number of elements $K$ per dictionary.

\textbf{Output}: Optimized dictionaries: $\{\mathbf{C}_m', m\in1\cdots M\}$

\begin{algorithmic}[1]
\STATE $\mathbf{C}_m'=\mathbf{C}_m, m=1\cdots M$
\REPEAT 

\STATE Arrange dictionaries in norm descending order: 
$$\sum\lVert\mathbf{c}_m'\rVert^2>\sum\lVert\mathbf{c}_{m+1}'\rVert^2, m\in 1\cdots M-1$$
\STATE Use beam search encoding method described in Section \ref{encoding} to encode $\mathbf{X}$:
$$\mathbf{x} = \sum_{m=1}^M \mathbf{c}_m'(i_m(\mathbf{x})) + \mathbf{e}_\mathbf{x}$$
where $\mathbf{e}_\mathbf{x}$ is the residue of $\mathbf{x}$.
\STATE Randomly pick a dictionary $\mathbf{C}_m'$, use the method described in Section \ref{learn}  to optimize the dictionary to better fit the intermediate dataset: $$\mathbf{X}'=\{\mathbf{x}'=\mathbf{e}_\mathbf{x}+\mathbf{c}_m'(i_m(\mathbf{x})), \mathbf{x}\in\mathbf{X}\}$$
(Firstly seek an $d_1$-dimensional subspace, where $d_1=d\cdot2^{S(\mathbf{C}_m)}/K$, then iteratively padding zeros and to fit higher dimensional subspace)
\UNTIL {Quit Condition}
\end{algorithmic}
\end{algorithm}

\subsection{Generate and fit the intermediate datasets}
As mentioned above, residual vector quantization generates largely mutually independent feature spaces, though traditional k-means method ended up with poor partitions. Anyway, the residual space is independent to all the dictionaries' feature space. So we add the residue to a dictionary's recovered dataset to generate an intermediate dataset:
$$\mathbf{X}'=\{\mathbf{x}'=\mathbf{e}_\mathbf{x}+\mathbf{c}_m'(i_m(\mathbf{x})), \mathbf{x}\in\mathbf{X}\}$$
and fit this new space to increase the independence of this dictionary as well as decrease the quantization error. If we find a dictionary fits the intermediate dataset better, the quantization error is lowered, and the independence of this dictionary is increased. Then the problem comes to how to learn a balanced partition, and to lower the quantization error.

\subsubsection{Subspace Clustering}
For the intermediate dataset, we seek a dictionary minimizing the residue as well as having high information entropy. An interesting observation on the residue is that it stultifies k-means algorithm throughly, as illustrated in Figure \ref{figInfo}, the information entropy could even drop to below 5-bits on the 7th dictionary of RVQ. 

To obtain a better clustering, one of the popular approaches is to cluster on lower-dimensional subspace \cite{agrawal1998automatic}, this is also what PQ/OPQ do to obtain high information entropy for each dictionary. Various previously proposed methods for high dimensional data clustering, e.g. \cite{bouveyron2007high}, \cite{jing2007entropy}, seek a clustering in an optimal subspace instead of the whole feature space. In lower-dimensional subspaces the projected datasets become denser and then a balanced clustering could be easily obtained. Also, clustering on subspaces could be more interpretive as irrelevant features could exist in high dimensional data. Some other approaches like PROCLUS \cite{bohm2004density}, uses a special distance function to assign each point to a unique cluster. 

However, in the case of fitting the intermediate dataset, it's not reasonable to clustering on just a few dimensions as the residue lies in the whole feature space. Also the distance function is already determined by applications. We seek a hybrid way to perform clustering on the intermediate dataset.

\subsubsection{Learning an Entropy Maximized Partition}
\label{learn}
We aim to optimize a dictionary instead of learn a new dictionary from scratch, as the dictionary learned previously could provide a better initial points for k-means\cite{bradley1998refining}. How much information of the dictionary should be used? If the dictionary fit the intermediate dataset well(like, have a high information entropy), then more information of the dictionary should be reserved. If the dictionary have a low information entropy, we should use reduce the dimension to initialize k-means on a small sub-space, so the noisy parts of the dictionary could be removed. Then we gradually adjust the dictionary to fit the whole feature space to obtain a more effective dictionary.

Here we suggest using $d_1=d\cdot2^{S(\mathbf{C}_m)}/K$ as the dimension of the subspace, as it directly measures if a dictionary is balanced. 
Following \cite{ding2004k}, we first perform PCA on the intermediate dataset and extract the component vectors: $\mathbf{R}=\{\mathbf{r}_{1}^T; \mathbf{r}_{2}^T; \cdots \mathbf{r}_{d}^T \}$. 
We then perform k-means on $\{\mathbf{R}_1\mathbf{x}'\}$, $\mathbf{R_1}=\{\mathbf{r}_{1}^T; \mathbf{r}_{2}^T; \cdots \mathbf{r}_{d_1}^T \}$, initialize it with dimension reduced dictionary $\{\mathbf{R}\mathbf{c}_m(k), k\in1\cdots K\}$. 
Iteratively, the learned dictionary(padded with zero chunks) is used to initialize the k-means on a bigger dimensional dataset: 
$\mathbf{R}_n\mathbf{x}'$, $\mathbf{R}_n=\{\mathbf{r}_1; \cdots \mathbf{r}_{d_n}^T ;\}, d_n>d_{n-1}$. Until we have learned the optimized dictionary $\{\mathbf{R}\mathbf{c}_m(k), k\in1\cdots K\}$ on $\{\mathbf{R}\mathbf{x}'\}$.

\subsection{Optimized Encoding}
\label{encoding}

\begin{table}
\centering
\caption{Comparison between different Encoding Schemes. GIST-1M dataset is used as it's very high-dimensional and tougher to obtain a better encoding. We randomly picked 1000 samples to perform the encoding experiments. We used dictionaries ($M=8,K=256$) optimized by DA initialized by RVQ. Our dictionary annealing (DA) encoding method and additive quantization (AQ) encoding method are compared. In addition, we implemented a "smart" brute force search which runs for hours encoding the vectors. We also used iterated conditional modes algorithm (ICM) to encode the dataset. DA and AQ are GPU accelerated by nVidia GTX980 with 4GB of dedicated memory, however, AQ's encoding scheme cannot fully utilize the GPU because it has more memory operations. ICM and brute force search is run on a Intel E5-2697v2 CPU with the latest Intel MKL.}
\begin{tabular}{|c|c|c|} \hline
Method     & Encoding Time & \shortstack{\\Quantization \\ Error}\\ \hline
DA($l=1$)  & 0.021s        & \shortstack{\\0.647480}\\ \hline
DA($l=10$) & 0.075s        & \shortstack{\\0.606554}\\ \hline
DA($l=100$)& 0.481s       & \shortstack{\\0.596206}\\ \hline
AQ($l=8$)  & 0.203s        & \shortstack{\\0.630149}\\ \hline
AQ($l=16$) & 0.259s        & \shortstack{\\0.619377}\\ \hline
AQ($l=32$) & 0.422s       & \shortstack{\\0.608681}\\ \hline
ICM\cite{besag1986statistical} & 26.504s& \shortstack{\\0.627182}\\ \hline
Brute Force & 10000s       & \shortstack{\\0.586213}\\ 
\hline\end{tabular}

\label{tabEncode}
\end{table}

Encoding for product quantization is quite simple since the original feature space has been divided into mutually orthogonal subspaces. For additive quantization\cite{barnes1993vector} and composite quantization\cite{cq}, the encoding problem is NP-hard. Encoding with the dictionaries optimized by DA is also NP-hard. For any input vector $\mathbf{x}$, we seek the code that minimize the quantization error $E$ :
\begin{equation}
\label{app}
\begin{split}
E&=\lVert \mathbf{x}-\sum_{m=1}^M \mathbf{c}_m(i_m(\mathbf{x}))\rVert^2 \\ &=\sum_{m=1}^M\lVert \mathbf{x}-\mathbf{c}_m(i_m(\mathbf{x}))\rVert^2-(m-1)\lVert \mathbf{x}\rVert^2 \\
&\quad+\sum_{a=1}^M\sum_{b=1,b\neq a}^M  {\mathbf{c}_a(i_a(\mathbf{x}))}^\mathrm{T}\mathbf{c}_b(i_b(\mathbf{x}))
\end{split}
\end{equation}
The third term above can be efficiently precomputed and stored for any input vector, and the second term can be omitted as it's a constant value. After that, the problem can be seen as a fully connected discrete pairwise MRF problem. The optimization of $E$ can be solved approximately by various existing algorithms. Additive quantization proposed a Beam Search algorithm in a matching pursuit fashion, the main idea is to maintain $L$ best approximations, and the overall time complexity encoding a input vector is $O(dMK+M^3KL\log L)$. Such encoding scheme could be very time consuming on large $M$. It also requires $L$ to be quite large to lower the quantization error as much as possible.
\newcommand\Xtilde{\stackrel{\sim}{\smash{\mathbf{x}}\rule{0pt}{1.1ex}}}
Suppose the best approximation (correct encoding) of a input vector is $\mathbf{x}\approxeq\mathbf{c}_1(i_1)+\mathbf{c}_2(i_2)+\cdots+\mathbf{c}_M(i_M)$. Further assume we have known the first $m-1$ correct encodings ${i_1, i_2,\cdots, i_{m-1}}$, can we effectively compute $i_m$? Denote the known part as $\hat{\mathbf{x}}=\mathbf{c}_1(i_1)+\cdots+\mathbf{c}_{m-1}(i_{m-1})$  and the unknown part as $\Xtilde=\mathbf{c}_1(i_{m+1})+\cdots+\mathbf{c}_{M}(i_{M})$, we seek the correct encoding on the m-th dictionary $i_m$. Notice that:

\begin{equation}
\begin{split}
\lVert\mathbf{x}-&\hat{\mathbf{x}}-\mathbf{c}_m(i_m)-\mathbf{x}'\rVert^2= \lVert\mathbf{x}-\hat{\mathbf{x}}\rVert^2 + \lVert\mathbf{x}-\Xtilde\rVert^2 + 2\hat{\mathbf{x}}^T\Xtilde\\
 & + \lVert\mathbf{x}-\mathbf{c}_m(i_m)\rVert^2 + 2\hat{\mathbf{x}}^T\mathbf{c}_m(i_m)+2\mathbf{c}_m(i_m))^T\Xtilde \\
 &-2\lVert\mathbf{x}\rVert^2
\end{split}
\end{equation}

The first three terms can be seen as a constant when we seek the correct $i_m$ , and the last term can be omitted. The fourth and fifth term can be effectively computed. However the sixth term cannot be computed because we don't know $\Xtilde$. If we omit this term extra error will be introduced. To lessen this error, we hope $\lVert\Xtilde\rVert$ is very small so that the variance of the last term won't have an serious impact on the final outcome. 

Thus we rearrange the dictionaries in the descending order of norm: $\sum\lVert\mathbf{c}_m'\rVert^2>\sum\lVert\mathbf{c}_{m+1}'\rVert^2, m\in 1\cdots M-1$. If our method is initialized with dictionaries learned with RVQ, the dictionaries naturally shrinks. We further adopted beam search on the scale shrinking dictionaries, that is, we maintain a list of best $L$ approximations of $\mathbf{x}$ on the first $(m-1)$ dictionaries: $\{\mathbf{a}_1^{m-1},\mathbf{a}_2^{m-1}, \cdots , \mathbf{a}_l^{m-1}\}$. Then we encode with the next dictionary $\mathbf{C}_m=\{\mathbf{c}_{m}(1), \mathbf{c}_{m}(2),\cdots,\mathbf{c}_{m}(K)\}$. We find $L$ combinations from $\{\mathbf{a}_{l}^{m-1}+\mathbf{c}_{m}(k)\}, l\in 1\cdots L, k\in 1\cdots K$ by minimizing the following objective function: 

\begin{equation}
\begin{split} 
\lVert \mathbf{x}-\mathbf{a}_{l}^{m-1}-\mathbf{c}_{m}(k)\rVert^2=&\lVert\mathbf{x}-\mathbf{a}_{l}^{m}\rVert^2+\lVert\mathbf{x}-\mathbf{c}_{m}(k)\rVert^2 \\
&-\lVert \mathbf{x} \rVert^2 + 2\mathbf{c}_{m}(k)^T\mathbf{a}_{l}^{m-1}
\end{split}
\end{equation}

The first term has been computed at the previous encoding step, and the third term $\lVert \mathbf{x} \rVert^2$ is constant for any $(\mathbf{a}_{l}^{m-1}+\mathbf{c}_{m}(k))$, thus negligible. And the last term involves $m$ table lookups and addition, with the inner-product of all dictionaries elements precomputed before the beam search procedure. Thus, only the term $\lVert\mathbf{x}-\mathbf{c}_{m}(k)\rVert^2$ is required to be computed. The time complexity is O($dK+MKL\log L$) for encoding with one single dictionary.

To sum up, our beam search iteratively uses the top $L$ candidates as seeds to find the best encoding for $\mathbf{x}$ with dictionaries arranged in a scale descending order. Our proposed method is quite similar to the multi-path search for residual tree \cite{kossentini1992large},  The overall time complexity is O($dMK+M^2KL\log L$). The encoding time grows with $M$. See Table \ref{tabEncode} for an empirical comparison with other encoding methods. It can be seen that at comparable quantization error, DA is much faster.

\section{Discussion about the Implementation Details}
\label{secDis}
\subsection{Initialization with different methods}

\begin{figure}[t]
\begin{center}
   \pgfplotsset{grid style={dotted}}
   	\begin{tikzpicture}
   		\begin{axis}[
   		width=0.4\textwidth,
   		height=0.25\textwidth,
   		axis y line=left,
   		axis x line=bottom,
   		ylabel=Quantization error,
   		ylabel near ticks,
   		xlabel=Iterations,
   		xmin=1, xmax=16,
   		ymin=0.55,ymax=0.75,
   		grid
   		]
   		\addplot[color=blue, mark=o] table [x=M, y=initPQ, col sep=comma] {figures/convergence.txt};

   		\addplot[color=red, mark=triangle] table [x=M, y=initRVQ, col sep=comma] {figures/convergence.txt};
   		
   		\addplot[color=magenta, mark=x] table [x=M, y=initRVQDA, col sep=comma] {figures/convergence.txt};
   		   		
   		\legend{PQ-DA, RVQ-DA, DARVQ-DA}   		
   		\end{axis}
   	
   	\end{tikzpicture}
\end{center}
   \caption{Convergence curve of dictionary annealing, initialized by dictionaries learned via product quantization and residual vector quantization methods on GIST-1M dataset. $M=8$ dictionaries are learned with $K=256$ elements per dictionary. For the encoding of DA, we use $L=10$. The vertical axis represents the quantization error and the horizontal axis corresponds to the number of iterations. The curves are Dictionary Annealing on dictionaries learned with Product Quantization, padded with zeros (PQ-DA), Dictionary Annealing on dictionaries learned with Residual Vector Quantization (RVQ-DA), and Dictionary Annealing on dictionaries learned with Dictionary Annealing Optimized Residual Vector Quantization (DARVQ-DA) }
\label{figConverge}
\end{figure}
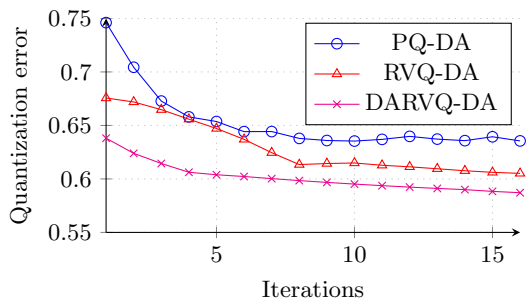
Our proposed method DA could be jointly used with other vector quantization: e.g. Additive Quantization, Product Quantization, Optimized Product Quantization, Composite Quantization, Residual Vector Quantization, etc. In addition, DA can be used right on the learning stages of RVQ: On each stage of RVQ, first use DA to optimize the dictionaries learned previously and encode the dataset, then perform k-means on the residue. We call this method Dictionary Annealing Optimized Residual Vector Quantization(DARVQ) in the following texts.

The selection of the initializing dictionary could have an impact on the convergence speed and the final outcome. We compared initializing dictionary annealing by PQ, by RVQ, and by DARVQ in Figure \ref{figConverge}, and we found DARVQ-DA has the lowest quantization error, followed by RVQ-DA. This is because residual vector quantization learned dictionaries' norm naturally reduces so the beam search in our proposed Dictionary Annealing method could perform much better. The norm of dictionaries learned by DARVQ shrinks even faster, so Dictionary Annealing could perform even better.

We can also observe that Dictionary Annealing reduces quantization error faster on the first $M$ iterations, that's because on the first $M$ iterations, the dictionaries are not balanced or not independent of each other. After first $M$ iterations, the dictionaries are balanced and mutually independent, so the improvement space is limited.

Here we suggest learning dictionaries with RVQ together with optimization by DA, and use these learned dictionaries to warm-start further optimization by DA. We used such initialization in all the following experiments.

\subsection{Scalability}

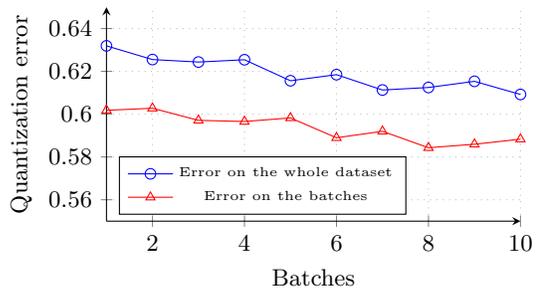
\begin{figure}[t]
\begin{center}
   \pgfplotsset{grid style={dotted}}\pgfplotsset{every axis/.append style={
                       legend style={font=\tiny},
                       }}
   	\begin{tikzpicture}
   		\begin{axis}[
   		width=0.4\textwidth,
   		height=0.25\textwidth,
   		axis y line=left,
   		axis x line=bottom,
   		ylabel=Quantization error,
   		ylabel near ticks,
   		xlabel=Batches,
   		xmin=1, xmax=10,
   		ymin=0.55,ymax=0.65,
   		legend pos=south west,
   		grid
   		]
   		\addplot[color=blue, mark=o] table [x=batch, y=overall, col sep=comma] {figures/online64.txt};

   		\addplot[color=red, mark=triangle] table [x=batch, y=fit, col sep=comma] {figures/online64.txt};
   		
   		\legend{Error on the whole dataset, Error on the batches}   		
   		\end{axis}
   	
   	\end{tikzpicture}
\end{center}
   \caption{Online training of dictionary annealing, initialized with dictionaries learned by Dictionary Annealing Optimized Residual Vector Quantization (DARVQ) on GIST-1M dataset. $M=8$ dictionaries are learned with $K=256$ elements per dictionary. We divided the whole one million data into 10 big batches to simulate the online training. The vertical axis represents the quantization error and the horizontal axis corresponds to the number of batches fed to DA.}
\label{figOnline}
\end{figure}
Dictionary Annealing can also be used for fitting online datasets. For a large-scale dataset, performing k-means on all data could be prohibitive, and the size of datasets may grow with time. Our proposed dictionary annealing can adjusts the dictionary to fit the new coming data. Dictionary Annealing gradually finds an optimal dictionary close to the original dictionary, instead of discarding all previously learned information.

The online dictionary learning is done simply by performing dictionary annealing on datasets in batches. We update the dictionary with large batches to prevent "misleading" the optimization. The overall quantization error of the dataset is further reduced by such online learning process, see Figure \ref{figOnline} for a demonstration. 

Online Dictionary Learning for sparse encoding has been proposed in \cite{mairal2009online}, while we focus on boosting the performance of ANN tasks. Our online dictionary learning scheme largely prevents searching performance from degrading on very large datasets, while it's very easily implemented. 

\subsection{Acceleration with the latest computation technologies}

\begin{table}
\centering
\caption{Computing time (in second) for different quantization methods on GIST-1M and SIFT-1M dataset, with $M=8, K=256$(64-bit encoding). All methods are GPU accelerated for fair. We used 100K samples for training, and encoded the whole dataset, finally we performed 1000 queries. DA(online) fitted the whole dataset in big bathes (100,000 samples per batch). DA encodes the dataset by maintaining $l=10$ best approximations, and for AQ encoding $l=32$. We can see by results that speeds of different methods could vary a lot. The degree of parallelism and cache friendliness impacts the speed of these algorithms as well as the time complexity.}
\begin{tabular}{|c|c|c|c|c|} \hline
Dataset                   & Method          & Training     & \shortstack{\\Encoding} & Query    \\ \hline
\multirow{6}{*}{GIST-1M}  & DA(online)      & 778.31s        & \shortstack{\\78.34s}     & 5.315s   \\ 
                          & DA(offline)     & 94.46s         & \shortstack{\\74.65s}     & 5.192s   \\ 
                          & AQ              & 414.69s        & \shortstack{\\392.33s}    & 5.224s   \\ 
                          & PQ              & 8.35s        & \shortstack{\\3.12s}    & 5.001s   \\ 
                          & OPQ             & 254.67s        & \shortstack{\\15.56s}    & 5.130s   \\ 
                          & RVQ             & 62.20s        & \shortstack{\\25.31s}    & 5.282s   \\
\hline
\multirow{6}{*}{SIFT-1M}  & DA(online)      & 527.46s        & \shortstack{\\67.87s}     & 5.231s   \\ 
                          & DA(offline)     & 64.56s        & \shortstack{\\64.56s}     & 5.282s   \\ 
                          & AQ              & 339.17s        & \shortstack{\\319.17s}    & 5.149s   \\ 
                          & PQ              & 5.46s        & \shortstack{\\1.73s}    & 4.993s   \\ 
                          & OPQ             & 95.84s        & \shortstack{\\2.48s}    & 5.162s   \\ 
                          & RVQ             & 22.19s        & \shortstack{\\11.51s}    & 5.295s   \\
\hline
\end{tabular}

\label{tabSpeed64b}
\end{table}

\begin{table}
\centering
\caption{Computing time for learning, encoding and searching with 128-bit encoding on SIFT-1M and GIST-1M datasets of different methods. }
\begin{tabular}{|c|c|c|c|c|} \hline
Dataset                   & Method          & Training     & \shortstack{\\Encoding} & Query    \\ \hline
\multirow{6}{*}{GIST-1M}  & DA(online)      & 3109.43s        & \shortstack{\\200.03s} & 9.479s   \\ 
                          & DA(offline)     & 379.15         & \shortstack{\\197.65s} & 9.415s   \\ 
                          & AQ              & 1225.02        & \shortstack{\\1131.96s}& 9.582s   \\ 
                          & PQ              & 20.35s        & \shortstack{\\3.45s}    & 9.218s   \\ 
                          & OPQ             & 333.26s        & \shortstack{\\15.08s}   & 9.408s   \\ 
                          & RVQ             & 116.13s        & \shortstack{\\40.99s}  & 9.563s   \\
\hline
\multirow{6}{*}{SIFT-1M}  & DA(online)      & 3242.32s        & \shortstack{\\178.18s} & 9.484s   \\ 
                          & DA(offline)     & 318.97s         & \shortstack{\\185.75s} & 9.475s   \\ 
                          & AQ              & 1206.88s        & \shortstack{\\1078.27s}  & 9.416s   \\ 
                          & PQ              & 10.74s        & \shortstack{\\1.97s}      & 9.270s   \\ 
                          & OPQ             & 176.81s        & \shortstack{\\2.66s}     & 9.300s   \\ 
                          & RVQ             & 43.20s        & \shortstack{\\25.31s}    & 9.432s   \\
\hline
\end{tabular}

\label{tabSpeed128b}
\end{table}

Our proposed dictionary annealing can be easily accelerated by the latest computation technologies. There is no branch in dictionary annealing, therefore implementation on GPU is quite easy with significant speed boost. 

For the dictionary optimization procedure, the major computation involves k-means and our proposed encoding scheme. K-means algorithm is very easily implemented on GPU \cite{farivar2008parallel}, multi-core system, and implement with the latest instruction sets such as AVX/AVX2\cite{wu2013vectorized}. For the encoding procedure. Our proposed encoding scheme requires enumerating $L$ best approximations of an input vector from a $KL$ combination lists, which requires intensive memory operations and less GPU-friendly. However, compared to the encoding method of Additive Quantization, our proposed encoding scheme requires less best approximations to be enumerated from a shorter list of total combinations. So our proposed encoding method is still ways faster.

We have implemented our dictionary annealing method with MATLAB, we have also used GPU acceleration, so the entire experiments below can be done rather fast. We also implemented other quantization methods on GPU. We reported the running time of experiments done in Section \ref{test} for different methods on Table \ref{tabSpeed64b} and Table \ref{tabSpeed128b}. On the dataset preparation, the majority of the time spent with DA is on the encoding stages, as well as AQ. Since our encoding method is faster than AQ, our approaches run much faster than AQ. Compared to OPQ, which has a significant speed loss on very high dimensions (mainly due to the time costly SVD decomposition), our proposed method can handle very high dimensional data easily. For the query time, though AQ, RVQ and our DA requires an additional fix to compute the approximated distance, it actually doesn't affect the query time: this is because on modern memory device it takes almost the same time to perform memory chunks copy or reset the memory chunks. The slight query time difference is due to pre-computation: AQ/RVQ/DA requires $O(MKd)$ time computing the distance between dictionaries elements and the query, while PQ/OPQ requires only $O(Kd)$. OPQ requires an additional vector rotation operation which takes $O(d^2)$ time.

\section{Performance on ANN tasks}
\label{test}

In this section we report the ANN tasks performance of dictionaries optimized by Dictionary Annealing, and compare it to the other state-of-the-art methods.

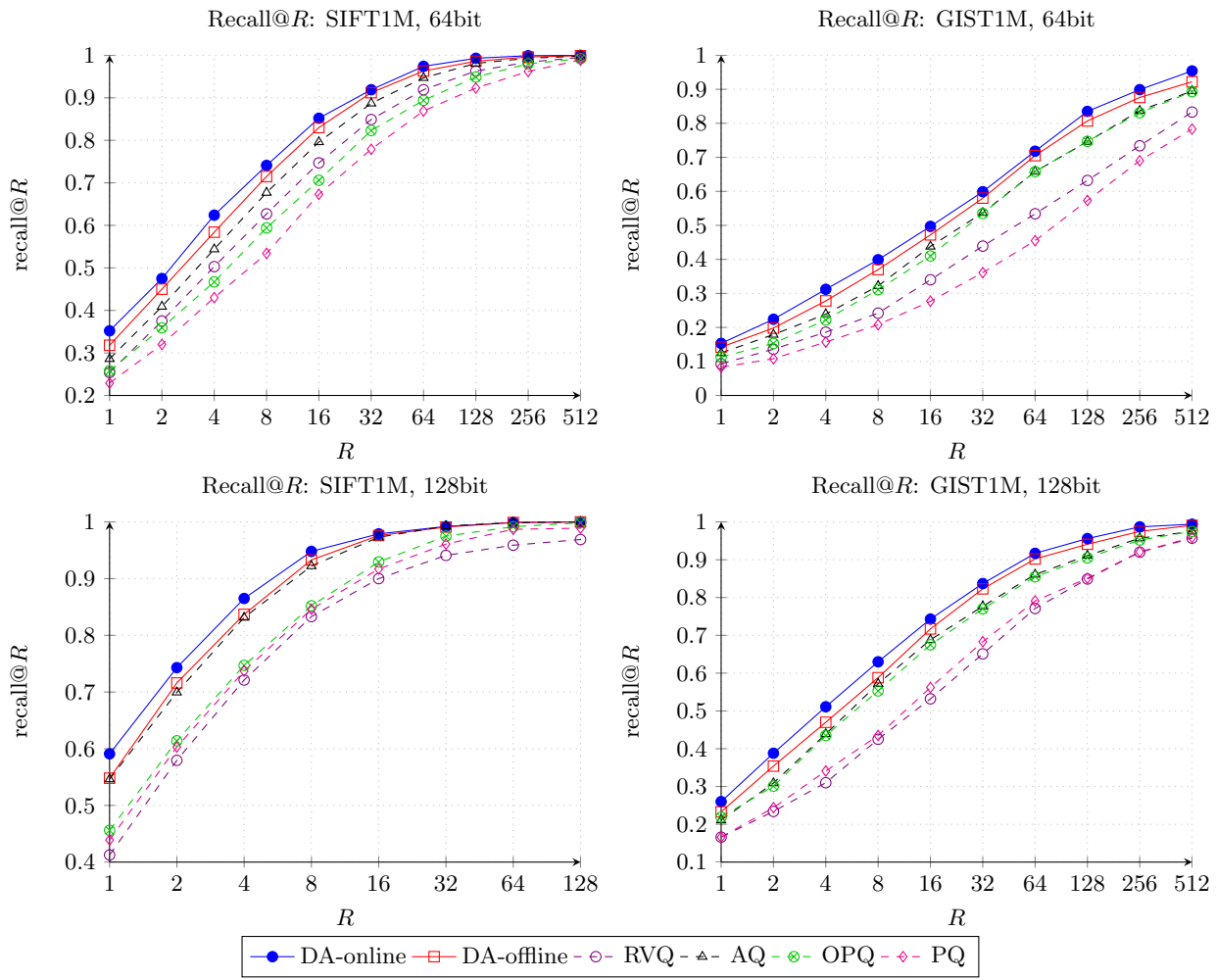
\begin{figure*}[t]
\begin{center}
   \pgfplotsset{grid style={dotted}}
   \pgfplotscreateplotcyclelist{my}{
   	{blue, mark=*},
   	{red,mark=square,mark options={solid}},
   	{dashed, violet,mark=o,mark options={solid}},
   	{dashed, black, mark=triangle,mark options={solid}},
   	{dashed, black!20!green, mark=otimes,mark options={solid}},
   	{dashed, magenta, mark=diamond,mark options={solid}},
   	{teal, mark=o},
   }
   	\begin{tikzpicture}
   		\begin{axis}[
   			title={Recall@$R$: SIFT1M, 64bit},
   			width=0.45\textwidth,
   			height=0.35\textwidth,
   			axis y line=left,
   			axis x line=bottom,
   			ylabel=recall@$R$,
   			xlabel=$R$,
   			xtick={1,2,4,8,16,32,64,128,256,512},
   			xticklabels={1,2,4,8,16,32,64,128,256,512},
   			ytick={0,0.1,0.2,0.3,0.4,0.5,0.6,0.7,0.8,0.9,1},   			
   			ymin=0.2, ymax=1,
   			xmode=log,
   			legend pos=south east,
   			grid,
   			cycle list name=my,
   			legend columns=-1,
   			legend entries={{DA-online},{DA-offline},{RVQ},{AQ},OPQ, PQ},
   			legend to name=leg64
   			]
   			
   			\addplot table [x=R, y=online, col sep=comma] {figures/recallSIFT1M64.txt};
   			
   			\addplot table [x=R, y=offline, col sep=comma] {figures/recallSIFT1M64.txt};

   			\addplot table [x=R, y=RVQ, col sep=comma] {figures/recallSIFT1M64.txt};
   			
    		\addplot table [x=R, y=AQ, col sep=comma] {figures/recallSIFT1M64.txt};  		
   			   			
   			\addplot table [x=R, y=OPQ, col sep=comma] {figures/recallSIFT1M64.txt};
   			
   			\addplot table [x=R, y=PQ, col sep=comma] {figures/recallSIFT1M64.txt};
   			
   		\end{axis}
   	
   	\end{tikzpicture}
   	\begin{tikzpicture}
   		\begin{axis}[
   			title={Recall@$R$: GIST1M, 64bit},
   			width=0.45\textwidth,
   			height=0.35\textwidth,
   			axis y line=left,
   			axis x line=bottom,
   			ylabel=recall@$R$,
   			xlabel=$R$,
   			xtick={1,2,4,8,16,32,64,128,256,512},
   			xticklabels={1,2,4,8,16,32,64,128,256,512},
   			xmode=log,
   			ytick={0,0.1,0.2,0.3,0.4,0.5,0.6,0.7,0.8,0.9,1},   			
   			ymin=0.0, ymax=1,
   			legend pos=south east,
   			grid,
   			cycle list name=my,
   			]
   			
   			
   			\addplot table [x=R, y=online, col sep=comma] {figures/recallGIST1M64.txt};
   			
   			\addplot table [x=R, y=offline, col sep=comma] {figures/recallGIST1M64.txt};

   			\addplot table [x=R, y=RVQ, col sep=comma] {figures/recallGIST1M64.txt};
   			
    		\addplot table [x=R, y=AQ, col sep=comma] {figures/recallGIST1M64.txt};  		
   			   			
   			\addplot table [x=R, y=OPQ, col sep=comma] {figures/recallGIST1M64.txt};
   			
   			\addplot table [x=R, y=PQ, col sep=comma] {figures/recallGIST1M64.txt};
   			
   		\end{axis}
   		
   		\end{tikzpicture}
   		
    	\begin{tikzpicture}
    		\begin{axis}[
    			title={Recall@$R$: SIFT1M, 128bit},
    			width=0.45\textwidth,
    			height=0.35\textwidth,
    			axis y line=left,
    			axis x line=bottom,
    			ylabel=recall@$R$,
    			xlabel=$R$,
    			xtick={1,2,4,8,16,32,64,128},
    			xticklabels={1,2,4,8,16,32,64,128,256,512},
    			ytick={0,0.1,0.2,0.3,0.4,0.5,0.6,0.7,0.8,0.9,1},
    			xmin=1,xmax=128,
    			ymin=0.4, ymax=1,
    			xmode=log,
    			legend pos=south east,
    			grid,
    			cycle list name=my,
    			legend columns=-1,
    			legend entries={{IO($l$=30,$I$=10)},{IO($l$=10,$I$=10)},{IO($l$=10,$I$=1)},{IO($l$=1,$I$=10)},Plain RVQ, PQ, CKM, CQ},
    			legend to name=leg128
    			]
    			\addplot table [x=R, y=online, col sep=comma] {figures/recallSIFT1M128.txt};
    			
    			\addplot table [x=R, y=offline, col sep=comma] {figures/recallSIFT1M128.txt};
 
    			\addplot table [x=R, y=RVQ, col sep=comma] {figures/recallSIFT1M128.txt};
    			
     		\addplot table [x=R, y=AQ, col sep=comma] {figures/recallSIFT1M128.txt};  		
    			   			
    			\addplot table [x=R, y=OPQ, col sep=comma] {figures/recallSIFT1M128.txt};
    			
    			\addplot table [x=R, y=PQ, col sep=comma] {figures/recallSIFT1M128.txt};
 
    		\end{axis}
    	
    	\end{tikzpicture}
    	\begin{tikzpicture}
    		\begin{axis}[
    			title={Recall@$R$: GIST1M, 128bit},
    			width=0.45\textwidth,
    			height=0.35\textwidth,
    			axis y line=left,
    			axis x line=bottom,
    			ylabel=recall@$R$,
    			xlabel=$R$,
    			xtick={1,2,4,8,16,32,64,128,256,512},
    			xticklabels={1,2,4,8,16,32,64,128,256,512},
    			xmode=log,
    			ytick={0,0.1,0.2,0.3,0.4,0.5,0.6,0.7,0.8,0.9,1},
    			ymin=0.1, ymax=1,
    			legend pos=south east,
    			grid,
    			cycle list name=my,
    			]
    			
    			
    			\addplot table [x=R, y=online, col sep=comma] {figures/recallGIST1M128.txt};
    			
    			\addplot table [x=R, y=offline, col sep=comma] {figures/recallGIST1M128.txt};
 
    			\addplot table [x=R, y=RVQ, col sep=comma] {figures/recallGIST1M128.txt};
    			
     		\addplot table [x=R, y=AQ, col sep=comma] {figures/recallGIST1M128.txt};  		
    			   			
    			\addplot table [x=R, y=OPQ, col sep=comma] {figures/recallGIST1M128.txt};
    			
    			\addplot table [x=R, y=PQ, col sep=comma] {figures/recallGIST1M128.txt};

    		\end{axis}
    		
    		\end{tikzpicture}   		
   		\ref{leg64}
\end{center}
   \caption{The performance for different algorithms on SIFT-1M and GIST-1M, with 64 bits encoding($M=8, K=256$).}
\label{fig64b}
\end{figure*}

\subsection{Datasets}

We performed the ANN search tests on the two datasets commonly used to validate the efficiency of ANN methods: SIFT-1M and GIST-1M from \cite{pq}:
\begin{description}
\item[SIFT1M] contains one million of 128-d SIFT \cite{sift} features. It's commonly used local feature descriptor for various image related applications.

\item[GIST1M] contains one million of 960-d GIST \cite{gist} global descriptors.
\end{description}
 
For each dataset, we randomly pick 100,000 vectors as the learning set. We then encode the rest of the database vectors, and perform 1000 queries to check ANN search quality. 

\subsection{Evaluated Methods}
 
We compared our DA to the following state-of-the-art quantization methods:
\begin{description}
\item[PQ]: Product quantization proposed in \cite{pq}. Following \cite{pq}, we used the structured ordering for GIST-1M and the natural ordering for SIFT-1M.
\item[OPQ]: Optimized Product Quantization proposed in \cite{opq}. We adopted the non-parametric version of OPQ. Cartesian k-means, the algorithm proposed in \cite{ck} shares a similar idea and have the same performance with OPQ.
\item[AQ]: Additive Quantization\cite{babenko2014additive}. Another similar algorithm is Composite Quantization\cite{cq}, which introduced a constraint named inner-dictionary-element-product on the encoding of the vectors to prevent computation of a "bias" in the asymmetric distance computation.
\item[RVQ]: Residual Vector Quantization proposed in \cite{rvq}.
\end{description}

For all the methods, we choose $k=256$ as the size of each dictionary, because it results a small look-up table and each subindex fits into one byte, which is instruction and cache friendly to modern CPU/GPUs. We choose $M=8/16$ to encode short codes for the dataset, resulting 64bit/128bit encodings. Such encoding greatly compressed the original dataset. For SIFT local descriptors, the original vector is 128d floating point numbers, which takes 128*32bits space. Quantization methods gain 1/64 compression ratio. For GIST global descriptors the compression ratio is even lowered to 1/480. An in-memory exhaustive search for these datasets is feasible.

For our DA methods, we conducted $M=8/16$ iterations. We used the dictionary obtained by DARVQ to initialize Dictionary Annealing. On optimizing with intermediate dataset, we let the dimensions grow exponentially to $d$ (the original dimensions of input dataset) in 5 iterations. In addition, we conducted the online dictionary optimization with Dictionary Annealing to learn a dictionary better fitting the whole dataset. Datasets are fed to DA in 100K sample batches.

To find ANNs, we perform linear scan search with asymmetric distances computation(ADC) proposed in \cite{pq}, which directly compare the input query and the quantized dataset. The search quality is measured using recall@$R$, which means for each query, we retrieved $R$ nearest items and check whether they contain the true nearest neighbor. Such criterion is commonly used to check efficiency of ANN methods.

Since the quantization based ANN search methods outperform hashing based binary embedding techniques \cite{opq}, \cite{pq}, \cite{ck}, we do not present the results of hashing performance in our tests.

\subsection{Results}

\begin{table}
\centering
\caption{Squared quantization error($E=\lVert\mathbf{x}-\hat{\mathbf{x}}\rVert^2$) on GIST-1M dataset of different quantization methods, with $M=8/16, K=256$. DA encodes with $l=10$, and AQ encodes with $l=32$.}
\begin{tabular}{|c|c|c|} \hline 
Method          & 64bit       & \shortstack{\\128bit}   \\ \hline
DA(online)      & 0.609222        & \shortstack{\\0.464948}     \\ \hline
DA(offline)     & 0.637022       & \shortstack{\\0.492671}     \\ \hline
AQ              & 0.679694       & \shortstack{\\0.521014}    \\ \hline
PQ              & 0.742063       & \shortstack{\\0.606044}     \\ \hline
OPQ             & 0.680419        & \shortstack{\\0.531976}    \\ \hline
RVQ             & 0.727788        & \shortstack{\\0.618995}    \\
\hline\end{tabular}

\label{tabQuantGIST}
\end{table}

\begin{table}
\centering
\caption{Squared quantization error on SIFT-1M dataset of different quantization methods, with $M=8/16, K=256$. DA encodes with $l=10$, and AQ encodes with $l=32$.}
\begin{tabular}{|c|c|c|} \hline 
Method          & 64bit       & \shortstack{\\128bit}   \\ \hline
DA(online)      & 16479.11        & \shortstack{\\7858.23}     \\ \hline
DA(offline)     & 17648.08       & \shortstack{\\9148.75}     \\ \hline
AQ              & 19032.97       & \shortstack{\\9176.82}    \\ \hline
PQ              & 23106.71       & \shortstack{\\10332.61}     \\ \hline
OPQ             & 21183.56        & \shortstack{\\9831.85}    \\ \hline
RVQ             & 20067.97        & \shortstack{\\9901.05}    \\
\hline\end{tabular}

\label{tabQuantSIFT}
\end{table}

Figure \ref{fig64b} shows the performance comparisons between different methods on 64bits and 128bit codes. One can see that our DARVQ-DA optimized dictionaries offer significant improvements to the original RVQ dictionaries. For example, on 64bit encoding, with DARVQ-DA we obtained 31.8\% recall@1 on SIFT-1M, while RVQ is only 25.4\%, the relative improvement is 25.2\%. The improvement gain is even larger on higher dimensional data GIST-1M, where we gained 14.6\% recall@1 with DARVQ-DA and only 9.3\% with RVQ, relatively 56.9\% improvement.

Using our DA optimized dictionaries for ANN tasks also outperforms other state-of-the-art methods. The offline DA optimized dictionaries with DARVQ outperforms Additive Quantization by 11.8\% , and the online version outperforms Additive Quantization by 16\% in terms of Recall@1 on 64bits encoding on SIFT1M. On 128bits encoding AQ and DA-offline performs similar, we speculate that AQ has already found near-optimal dictionaries fitting the learning dataset so the improvement is limited(DA-online learns a better dictionary with all the data). Generally DA optimized dictionaries has the best performance with noticeable advantage. That's because our Dictionary Annealing could gain a lower quantization error. The quantization error of different methods are reported on Table \ref{tabQuantGIST} and Table \ref{tabQuantSIFT}.


\section{Conclusion and future works}
In this paper, we introduced Dictionary Annealing method for optimizing dictionaries used by quantization based approximate nearest neighbor search methods. We first discussed what makes good encoding: high inter-dictionary independence and high inner-dictionary information entropy. We observed that residual vector quantization easily produces independent dictionaries, and clustering on subspace generates a balanced partition. Motivated by these observations, we use residual vectors to increase the independence of dictionaries, and perform warm-started k-means with clusters on subspaces to learn better dictionaries. We also used an optimized multi-path encoding method to aid the dictionary annealing procedure. Dictionary Annealing could make significant improvements to the dictionaries learned by other methods, especially the dictionary learned by residual vector quantization. Empirical results on the SIFT-1M and GIST-1M datasets commonly used for evaluating ANN search methods demonstrated that our proposed approach outperforms existing methods. 

Our major contribution is to show optimizing dictionaries with residue could bring significant performance gain while not modifying the original framework intensively, and online optimizing dictionaries could bring even more performance gains. Currently, the main limitation of the proposed scheme is the speed of encoding. For more dictionaries our proposed method have to deal with growing inner-product variances of inter-dictionary elements. Our future work will focus on eliminating such variance, so further performance gains could be possible.


\bibliographystyle{abbrv}
\bibliography{sigproc}  
%
%

\end{document}